\documentclass[journal]{IEEEtran}
\usepackage{cite}
\ifCLASSINFOpdf
  \usepackage[pdftex]{graphicx}
  \graphicspath{ {./} }
\else
\fi
\usepackage{amssymb}
\usepackage{amsmath}
\usepackage{pifont}
\usepackage{bm}
\usepackage{siunitx}
\usepackage{tablefootnote}
\usepackage{algorithm}
\usepackage{algpseudocode}
\usepackage{booktabs}
\usepackage{array}
\newcolumntype{P}[1]{>{\raggedright\arraybackslash}p{#1}}
\usepackage{float}
\usepackage[hyphens]{url}
\usepackage{hyperref}
\usepackage{multicol}

\begin{document}
\title{Federated Learning for Short-term Residential Load Forecasting}

\author{Christopher~Briggs,
        Zhong~Fan,~\IEEEmembership{Senior~Member,~IEEE,}
        and~Peter~Andras,~\IEEEmembership{Senior~Member,~IEEE}% <-this % stops a space
\thanks{C. Briggs, Z. Fan and P. Andras are with the School of Computing and Mathematics, Keele University, Staffordshire, ST5 5BG, UK}% <-this % stops a space
% \thanks{Manuscript received April 19, 2005; revised August 26, 2015.}
}

\maketitle

\begin{abstract}
  Load forecasting is an essential task performed within the energy industry to help balance supply with demand and maintain a stable load on the electricity grid. As supply transitions towards less reliable renewable energy generation, smart meters will prove a vital component to facilitate these forecasting tasks. However, smart meter adoption is low among privacy-conscious consumers that fear intrusion upon their fine-grained consumption data. In this work we propose and explore a federated learning (FL) based approach for training forecasting models in a distributed, collaborative manner whilst retaining the privacy of the underlying data. We compare two approaches: FL, and a clustered variant, FL+HC against a non-private, centralised learning approach and a fully private, localised learning approach. Within these approaches, we measure model performance using RMSE and computational efficiency. In addition, we suggest the FL strategies are followed by a personalisation step and show that model performance can be improved by doing so. We show that FL+HC followed by personalisation can achieve a $\sim$5\% improvement in model performance with a $\sim$10x reduction in computation compared to localised learning. Finally we provide advice on private aggregation of predictions for building a private end-to-end load forecasting application.
\end{abstract}

\begin{IEEEkeywords}
federated learning, load forecasting, distributed machine learning, deep learning, data privacy, internet-of-things
\end{IEEEkeywords}

% \fi
\IEEEpeerreviewmaketitle

% \section{Introduction}
% for IEEE journal papers produced under \LaTeX\ using
% IEEEtran.cls version 1.8b and later.
% I wish you the best of success.

 

% \subsection{Subsection Heading Here}
% Subsection text here.

%\IEEEpubidadjcol

% \subsubsection{Subsubsection Heading Here}
% Subsubsection text here.

\section{Introduction}
\label{sec:introduction}

Smart meters are being deployed in many countries across the world for the purpose of optimising efficiency within electricity grids and providing consumers with insights into their energy usage. The meters record energy consumption within a building directly from the electricity supply and periodically communicate this data to energy suppliers and other entities in the energy sector. Smart meter data contain an enormous amount of potential predictive power that will aid the transition from fossil fuel technologies to cleaner and renewable technologies \cite{Anonymous:2019up}. However this high-resolution data is particularly sensitive as it can easily enable inference about household occupancy, lifestyle habits or even what and when specific appliances are being used in a household \cite{McKenna:2012ga}.

A large contribution of renewables in the energy mix poses a significant challenge for balancing supply and demand. If peak demand coincides with low wind/solar inputs, energy must be provided by reliable backup generation, such as idling gas turbines. Such solutions are very costly, both economically and environmentally and serve to discourage the installation of large amounts of renewable energy generation. Reliable forecasting will provide opportunity for more efficient optimisation of electricity grids to cope with varying energy demand.

Despite the benefits for promoting a greener energy sector, smart meter installation in most countries is an opt-in process and levels of adoption of smart meters are beginning to stagnate. Data privacy and security concerns are among the most cited reasons consumers give for rejecting a smart meter installation \cite{BaltaOzkan:2014dy}. Specific privacy concerns with smart meters include government surveillance, energy companies selling data and illegal data acquisition/use \cite{McKenna:2012ga}.

Deep learning \cite{2015Natur.521..436L} - a subset of machine learning that makes use of multi-layered neural networks for classification and regression tasks, among others - has shown great promise in many applications in recent years. Time-series forecasting is one such strength of deep learning \cite{2018arXiv181206127L} using specific architectures such as recurrent neural networks (RNNs) that are designed to capture temporal dependencies during training. Among the most successful RNN architectures for forecasting are Long-Short Term Memory networks (LSTMs), that learn long-range dependencies particularly well.

In this paper, we propose the use of a modern distributed machine learning setting known as federated learning (FL) \cite{mcmahan2017communication} to train load forecasting LSTM models while preserving the privacy of consumer energy consumption data that could enable greater adoption of smart meters by privacy-conscious consumers. Our main contributions are: (a) a thorough comparison of how FL training strategies and non-FL benchmarks affect a model's forecasting performance (b) a comparative analysis of FL to a FL variant, designed specifically to perform well over non-iid data, applied to load forecasting, (c) an evaluation of computational efficiency issues arising in the FL forecasting system, and (d) the identification of the necessity of a personalisation step in FL-based forecasting to improve model performance beyond training individual local models in isolation.

The remainder of this paper is organised as follows. A short literature study is presented in \autoref{sec:literature-review}. We explore properties of the smart meter energy demand dataset used in our experiments in \autoref{sec:exploratory-data-analysis}. In \autoref{sec:methodology} we provide our methodology and in \autoref{sec:results-and-discussion} we present our results along with discussion. Finally we conclude our work in \autoref{sec:conclusion}.

\section{Literature review}
\label{sec:literature-review}

In the literature, AI and machine learning have been adopted for load forecasting since 1990s. In particular, artificial neural networks have been the most popular technique during the past three decades \cite{9761176}, while fuzzy logic and support vector machines (SVM) have also been used in many papers \cite{1350819,5393627}. Since 2015 there has been a huge increase of applying deep learning to load forecasting, e.g. \cite{en11071636,7885096,8001465} Notably, the authors of \cite{7885096} have developed a bespoke deep learning application for household load forecasting and the method was tested on 920 smart metered customers from Ireland. It is shown to outperform some of the state-of-the-art techniques in household load forecasting, such as ARIMA (AutoRegressive Integrated Moving Average) and SVR (support vector regression) in terms of RMSE (Root Mean Square Error).

Despite the above technical advancements, as pointed out by the authors of \cite{9761176}, load forecasting is still an evolving field, and "no technique is superior to all other methods in load forecasting". Therefore, power system academics should work together with industry as well as researchers in other disciplines, such as big data, computer science, and meteorology, to facilitate wide deployment of better load forecasting models in practice.

 The most successful neural network architectures for forecasting are based on recurrent neural networks (RNNs), such as Long-Short Term Memory networks (LSTMs) \cite{Hochreiter:1997fq}. These architectures can learn what long and short term information to pay attention to during the training process. Recent surveys compare and contrast traditional and modern approaches to load forecasting and conclude that AI-based methods (such as those that utilise neural networks) offer the greatest predictive performance across all forecasting horizons \cite{Wei:2019ez,Bourdeau:2019fj}. Kong et al. \cite{Kong:cn} investigate the use of an LSTM architecture to predict short-term electrical load for residential properties. The authors show that forecasting with an LSTM outperforms other statistical and machine learning methods for this purpose. We draw inspiration from this work to form our comparative centralised learning approach and thus the architecture for our FL training scenarios.

The key drawback to how both traditional and AI-based methods have been applied in the load forecasting literature is the need for data to be centralised. Clearly the privacy of consumer energy consumption data can easily be violated in such cases. FL provides a key mechanism to tackle the issue of training a model over private data. FL research has its roots in distributed optimisation within the datacentre to deal with very large datasets \cite{Dean:2012wx}. The term `federated learning' was coined in a paper by researchers at Google who presented a simple distributed stochastic gradient descent (SGD) procedure known as federated averaging \cite{mcmahan2017communication} which allows a selection of devices to train on local data and contribute updates to a shared, global model. The procedure keeps raw data private but requires significantly greater wall-clock time to train models that can compete with models trained in the more conventional centralised fashion. One key concern with training models under FL is degraded optimisation performance and/or reduced model performance in the presence of non-IID data \cite{hsieh2020non}. Several approaches have been suggested to tackle this issue. One idea is to regularise the updates from individual devices to constrain the distance between local models and the global model \cite{2018arXiv181206127L}. Another approach is to abandon the idea of training a single global model in favour of multiple specialised models to fit divergent data. Such ideas include federated multi-task learning \cite{2017arXiv170510467S}  and clustered FL \cite{Sattler:2019tb}. In this paper we explore the effect of a variant of FL using hierarchical clustering (HC) known as FL+HC \cite{Briggs:it}, that introduces a hierarchical clustering algorithm during the FL procedure to partition devices by update similarity.

For load forecasting applications, few works exist that consider the use FL. The authors in \cite{10.1145/3396851.3397717} investigate how to predict chiller efficiency in HVAC systems with the goal of reducing energy consumption. The work compares a centralised learning approach with FL, concluding that FL model performance suffers when training over all installation sites but can be improved when data is grouped by installation site. In \cite{Saputra:2019ha}, the authors apply FL to predict energy demand in the scenario of electric vehicle charging networks. They show that clustering charging stations geographically prior to learning improved model performance and reduced communication overhead.

On smart meter data, \cite{9380668} apply FL to privately predict the value of various socio-demographic data features of each household in order for energy utilities to offer diversified services to their consumers. The work most similar to our own in \cite{9148937} provides a simple study of Fl for load forecasting using household energy consumption data. Where our work differs is the depth of our analysis and our comparison of training strategies including multiple model approaches to tackle the known issue of poor FL performance on non-IID datasets. We additionally benchmark against both centralised learning and localised learning - the latter already provides a fully-private forecast, so is very important to compare an FL system against. Finally we test a wide variety of time-series sequence lengths to understand how this affects learning in all our different approaches.
\section{Exploratory Data Analysis}
\label{sec:exploratory-data-analysis}
In order to test an application for short-term energy load forecasting, a suitable dataset with reliable real-world high to medium resolution electricity meter readings was required. Additionally, summarising and visualising the data and distribution of various facets of a dataset will allow us to draw insights about individual households' energy demand over time. This section briefly describes the dataset used, our sub-sample of the dataset and provides some exploratory visualisations of the sampled data.

\subsection{Dataset}
\label{sec:dataset}
The dataset used for our experiments was gathered under the Low Carbon London project delivered by UK Power networks \cite{Anonymous:2015wr}. This 4 year project was designed to support low carbon energy solutions within the UK and was conducted between 2011 and 2014. The project made available the smart electricity meter readings for a sample of 5,567 London households, many of which cover 1 or more years of the duration of the project. The data is provided as discretised 30-minute meter readings showing total energy consumption (in kWh) recorded within each interval.

In order to carry out a detailed comparative study between different training methods, a small sample of 100 households was randomly selected over the period 1st Jan 2013 to 30th June 2013. The selection criteria for these 100 households required that meter readings should cover the period described above and that the meters were gathering consumption data under a standard flat-rate electricity billing tariff (as opposed to a dynamic time of use tariff that was also present in the dataset). This final criterion was applied to reduce the behavioural bias that time of use tariffs induce in energy consumption habits within a household. The resulting sample was therefore expected to contain households who use energy with no influence other than their normal daily habits and occupancy.

In conjunction with the energy consumption data, we also considered how weather related data might impact on forecasting models trained under different scenarios. As all the consumption data is collected within the greater London area, it was possible to collect weather readings that could be easily fused with the consumption data. These included the air temperature (in degrees Celsius) and relative humidity (as a percentage) recorded by the Met Office \cite{UKhourlyweatherob:2020ut}. As the exact location of each household is not recorded in the dataset, the London Heathrow weather station was selected as it contained a full set of hourly readings for the duration of the study period.

A detailed description of all specific data pre-processing techniques that were applied to the resulting data sample in our study are provided in \autoref{sec:dataset-prep}

\subsection{Data visualisation}
\begin{figure}[t]
    \centering
    \includegraphics[width=\linewidth]{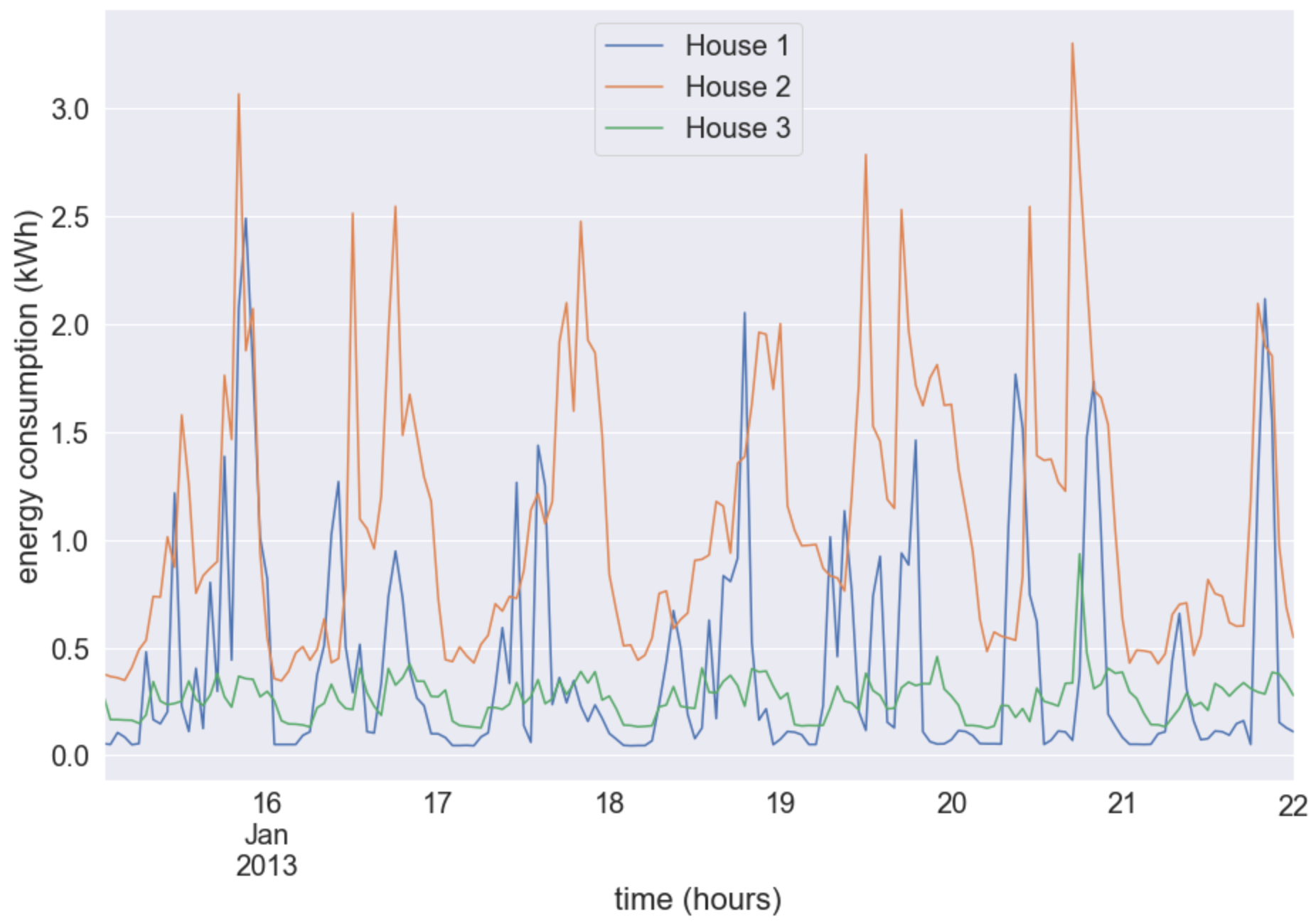}
    \caption{An example of the hourly energy consumption profiles for 3 random households in the sampled dataset over a 7 day period between 15th January 2013 and 22nd January 2013}
    \label{fig:example-hourly-energy-consumption}
\end{figure}

\begin{figure}[t]
    \centering
    \includegraphics[width=\linewidth]{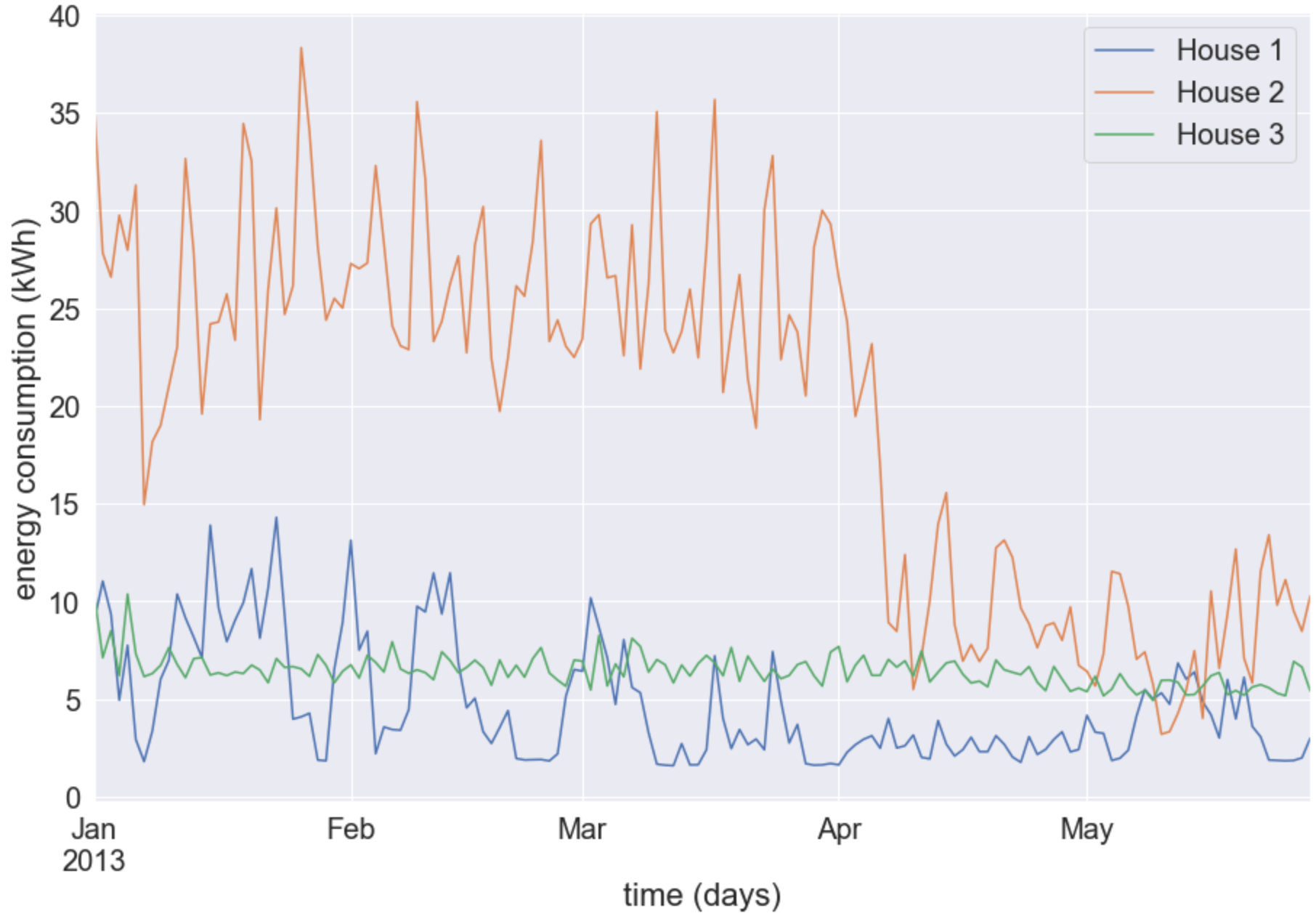}
    \caption{An example of the daily energy consumption profiles for the same 3 random households as in \autoref{fig:example-hourly-energy-consumption} in the sampled dataset over a 6 month period between 1st January 2013 and 30th June 2013}
    \label{fig:example-daily-energy-consumption}
\end{figure}

The hourly and daily energy consumption profiles of 3 random households from our sampled dataset (over 7 days and 6 month respectively) are presented in \autoref{fig:example-hourly-energy-consumption} and \autoref{fig:example-daily-energy-consumption}. From the hourly profiles, each household uses more energy during the day than at night as would be expected for most people. However, the maximum level of energy consumption is quite different among the households, as is the time of day when most energy is used. Houses 1 and 2 show 2 or 3 peaks roughly corresponding with increased energy consumption in the morning and evening, whereas House 3 uses energy more consistently throughout the day. Another insight that becomes clear from visualising the hourly data is that private habitual activity is visible at this granularity. For example, low energy consumption in House 1 on the evening of the 17th might suggest low or zero occupancy at that time, especially considering high energy consumption in the evenings of all other days in this time window.

Visualising the daily energy consumption profiles reveals that longer term energy usage is quite different over these same 3 households as well. House 1 uses more energy for 7-14 day periods followed by lower energy use. House 2 used more or less energy sporadically day to day with a considerable drop in energy use in early April (perhaps indicating electric heating use in the colder months which would account for the relatively high daily energy consumption). Finally, House 3 is incredibly consistent in its energy usage habits at this granularity, as was the case at the hourly resolution.

\begin{figure}[t]
    \centering
    \includegraphics[width=\linewidth]{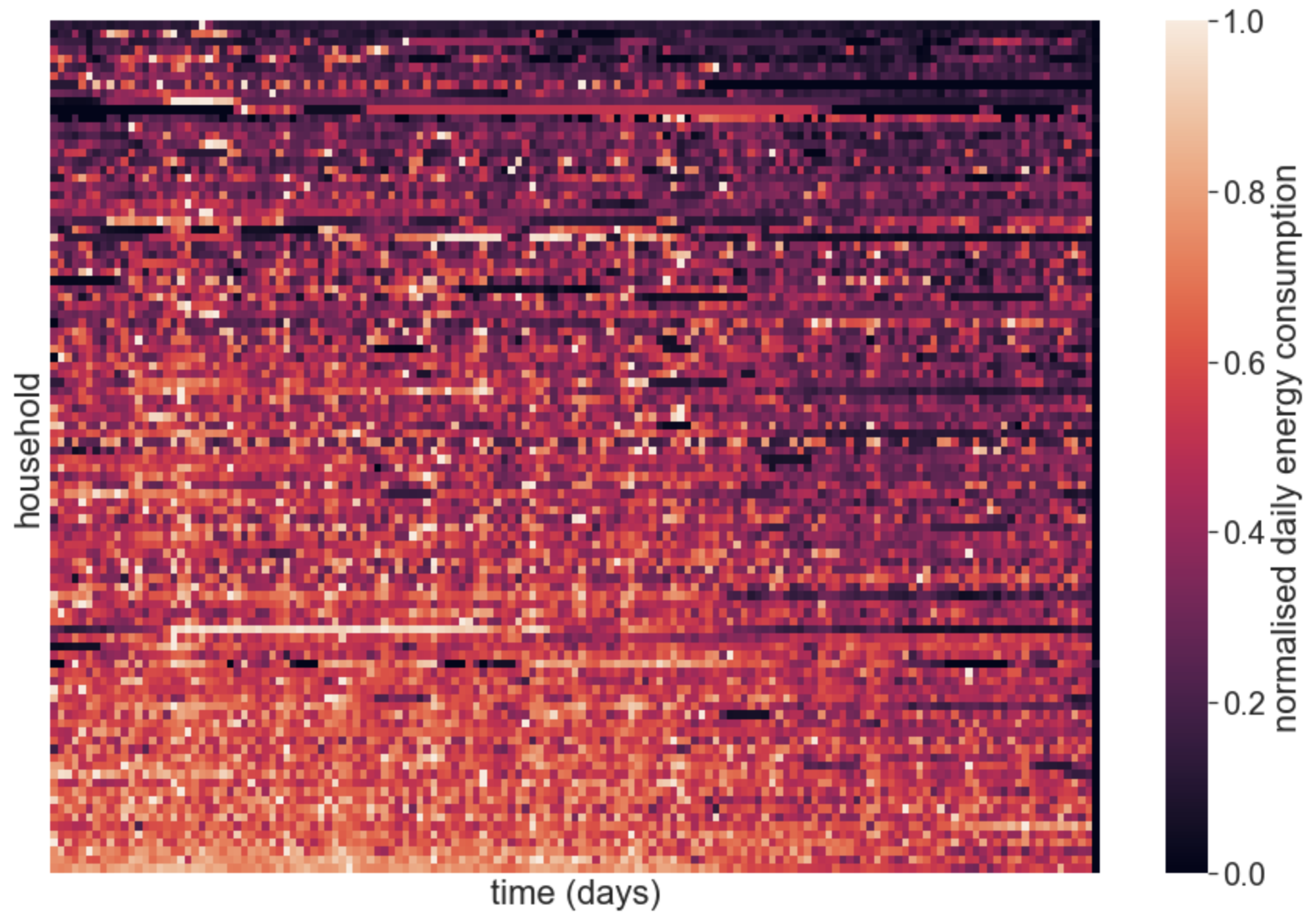}
    \caption{Aggregated daily energy consumption per household in the sampled dataset. The heatmap presents energy consumption as min-max normalised values (by household) between 0 and 1. Households are sorted (top to bottom) by total energy consumption.}
    \label{fig:energy-consumption-by-day}
\end{figure}

\begin{figure}[t]
    \centering
    \includegraphics[width=\linewidth]{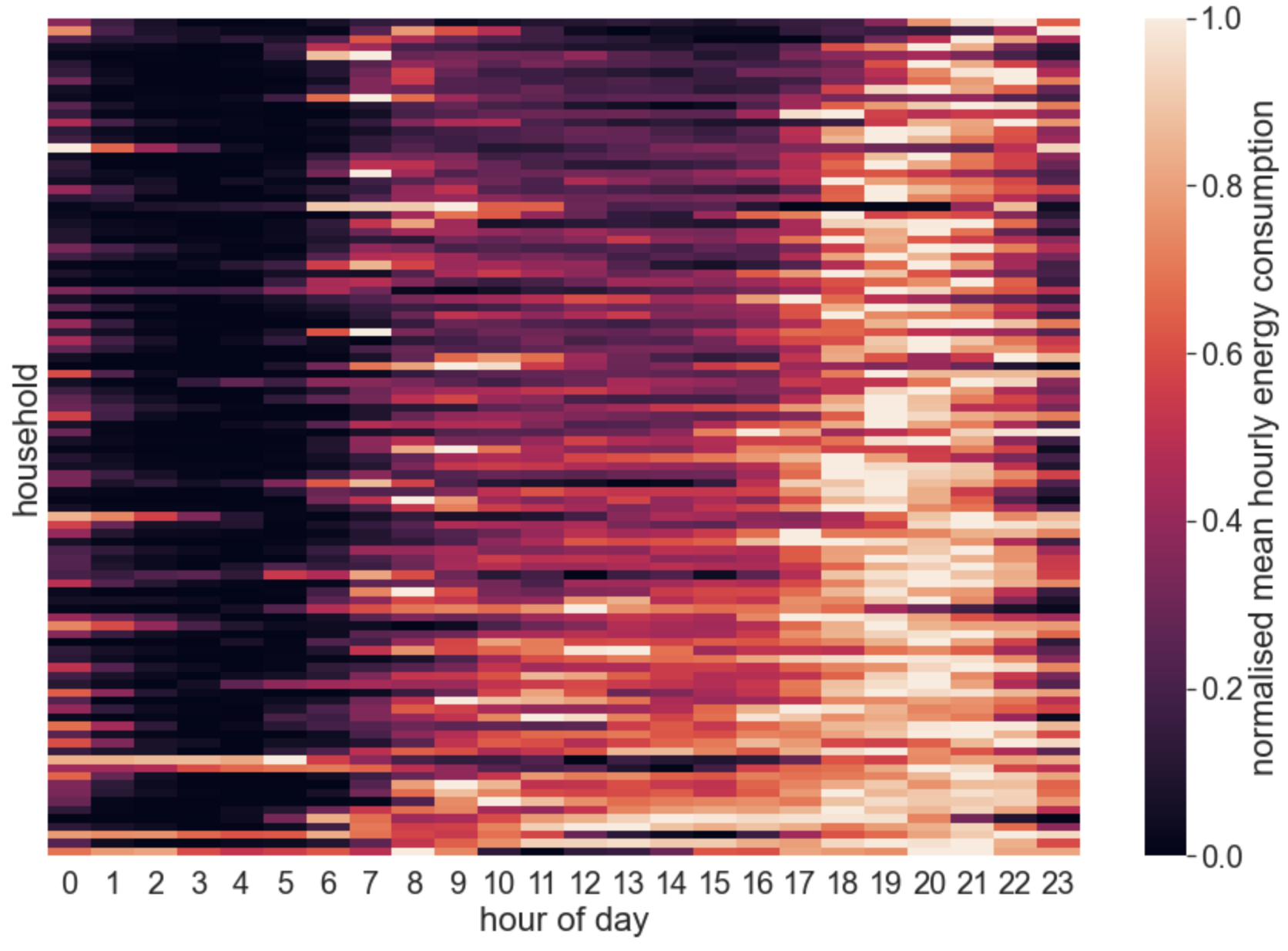}
    \caption{Mean hourly energy consumption per household in the sampled dataset. The heatmap presents energy consumption as min-max normalised values (by household) between 0 and 1. Households are sorted (top to bottom) by total energy consumption.}
    \label{fig:energy-consumption-by-hour}
\end{figure}

Visualising just 3 households from the sample reveals the non-iid nature of individual household energy consumption at both a high and low resolution. Clearly, any forecasting model built on this data will need to capture this variability in energy usage between households. More broadly, we have produced heatmaps showing the aggregated daily energy consumption per household (\autoref{fig:energy-consumption-by-day}) and the mean energy consumption by hour of the day for each houshold (\autoref{fig:energy-consumption-by-hour}). Both plots show min-max normalised energy consumption profiles (normalisation applied to each household individually) and are sorted by total energy consumption for each household.

\autoref{fig:energy-consumption-by-day} reveals a large variance in aggregated daily energy consumption between households. Additionally, households where energy consumption remains consistent day to day are visible, in contrast with households that display an irregular distribution of energy usage depending on the day. \autoref{fig:energy-consumption-by-hour} shows that peak energy demand tends to occur between 7am and 9am and 5pm until 10pm, likely consistent with occupancy and waking hours. However, the hour of peak energy demand shifts slightly from household to household which may prove difficult for a single joint forecasting model to represent. In this paper we will investigate how well several machine learning training approaches affect the ability of a model to produce accurate forecasts under the described non-iid data distribution over individual households' energy consumption habits.
\section{Methodology}
\label{sec:methodology}

\subsection{Dataset preparation}
\label{sec:dataset-prep}
After selecting 100 households from the Low Carbon London dataset (see \autoref{sec:dataset} for selection criteria), the raw energy consumption readings were passed though a pipeline of transformations to clean the data. For each individual household, this included dropping duplicated readings, forward filling empty readings and resampling the reading intervals to give an hourly record of energy consumption. A design matrix $\bm{X}_c$ was then built for each household $c$ based on this transformed data containing feature vectors of the form $\bm{x}_t = \{e_t, y_t, w_t, d_t, h_t\}$ for each time index $t$ composed of: 

\begin{enumerate}
    \item the energy consumption value $e_t$ in kWh
    \item the year $y_t$ corresponding with the time index
    \item the week of the year $w_t$ in the range 0-51
    \item the day of the week $d_t$ in the range 0-6
    \item the hour of the day $h_t$ in the range 0-23
\end{enumerate}

A second design matrix $\bm{W_c}$ was also built for each household $c$ that included weather data from the Met Office (discussed in more detail in \autoref{sec:dataset}). The feature vectors of the form $\bm{w}_t = \{e_t, y_t, w_t, d_t, h_t, a_t, r_t\}$ comprising this design matrix were additionally composed of:

\begin{enumerate}
    \item the recorded air temperature $a_t$ in degrees Celsius corresponding with the time index
    \item the calculated relative humidity $r_t$ (as a percentage)
\end{enumerate}

Finally both $\bm{X_c}$ and $\bm{W_c}$ for each household $c$ were partitioned into training, validation and testing datasets according to 0.7/0.2/0.1 split. As the time series data is sequential by its very nature, the validation split contains time indices strictly greater than those in the training split and the test split contains time indices strictly greater than those in the validation split.

\subsection{Forecasting task}
The forecasting task is designed specifically for how LSTMs ingest data to be trained to perform predictions. As such, the initial design matrices for each household are transformed into consecutive rolling sequences of $K$ feature vectors. For example, each sequence $\bm{S}_t \in \bm{S}$ drawn from the design matrix $\bm{X}_c$ takes the form $\bm{S}_t = \{\bm{x}_{t-K}, ..., \bm{x}_{t-2}, \bm{x}_{t-1}\}$. This forms a single input sequence to the LSTM. The corresponding label for this sequence is the energy consumption $e_t$ at time index $t$. The task of the LSTM is therefore to learn an appropriate mapping from $\bm{S} \rightarrow \hat{e}_t$ by minimising the error between the observed energy consumption $e_t$ and the predicted or forecasted energy consumption $\hat{e}_t$. For all our experiments we report the root mean squared error (RMSE) to compare the different training strategies set out in this paper:

\begin{equation}
    \text{RMSE} = \sqrt{\sum{(\hat{e}_t - e_t)^2}/N}
\end{equation}

We chose to create sequence datasets for $K=6$, $K=12$ and $K=24$ hours. As we started with two design matrices (with and without weather data fused), this results in 6 sequence datasets for each household which we denote:

\begin{multicols}{2}
\begin{itemize}
    \item $\bm{S}_{K=6,\text{+weather}}$
    \item $\bm{S}_{K=12,\text{+weather}}$
    \item $\bm{S}_{K=24,\text{+weather}}$
    \item $\bm{S}_{K=6,\text{-weather}}$
    \item $\bm{S}_{K=12,\text{-weather}}$
    \item $\bm{S}_{K=24,\text{-weather}}$
\end{itemize}
\end{multicols}

The forecasting task can be formally verbalised as: ``Predict the current energy consumption from the preceding $K$ hour's energy consumption readings''.

As LSTMs are more efficiently optimised when the data in different dimensions are equally scaled, we apply a min-max normalisation (independently in each dimension) to the data to ensure all values fall between 0 and 1. The minimum and maximum values for $e_t$ are drawn globally from across all the household datasets and therefore each sequence dataset $\bm{S}$ makes use of the same normalisation operation across all households.

\subsection{LSTM framework}
The long short term memory (LTSM) \cite{Hochreiter:1997fq} network belongs to a family of neural network architectures known as recurrent neural networks (RNNs). Such networks are designed to handle sequential data such as time series data or language fragments such as sentences. The major distinction between RNNs and standard feed forward neural networks is the ability to pass the output of hidden units back into themselves as well as incorporating gates to control the flow of past and current information. These conditions allow for learning of temporal patterns. For an energy forecasting problem a RNN can potentially learn how daily patterns of energy consumption affect future consumption by way of the memory built into RNNs.

The LSTM is one of the most sophisticated RNN architectures in that it works exceptionally well to store long-term temporal dependencies. Earlier RNN architectural designs are plagued with issues related to vanishing or exploding gradients during training via backpropagation \cite{279181}. Such issues resulted in the network becoming unable to learn anything from information earlier in the sequence beyond the preceding few time steps. LSTMs introduce an internal memory state that can persist over many time steps allowing the network to learn from long-term patterns.

In each LSTM cell, an internal state $\bm{c}_t$ is regulated by a forget gate $\bm{f}_t$ controlling the weight of information from the output during the previous time step $\bm{h}_{t-1}$ and the input for the current time step $\bm{x}_t$. The input feature for the current time step $\bm{i}_t$ is accumulated into the internal state under the influence of the input gate $\bm{g}_t$. Finally the output gate $\bm{o}_t$ governs the output $\bm{h}_t$ formed from the inputs and the internal cell state. The new internal state $\bm{c}_t$ and cell output $\bm{h}_t$ become inputs for the the cell at the next time step (additionally the final cell output $\bm{h}_t$ is passed to the next layer in a deep network). The memory cell state $\bm{c}_t$ and output activation $\bm{h}_t$ are calculated using equations \ref{eq:lstm-forget-gate} to \ref{eq:lstm-activation}.

\begin{align}
\begin{split}
    \label{eq:lstm-forget-gate}
    \bm{f}_t = \sigma(\bm{W}_{fx} \bm{x}_t + \bm{W}_{fh} \bm{h}_{t-1} + \bm{b}_f)
\end{split}\\
\begin{split}
    \label{eq:lstm-input-gate}
    \bm{i}_t = \sigma(\bm{W}_{ix} \bm{x}_t + \bm{W}_{ih} \bm{h}_{t-1} + \bm{b}_i)
\end{split}\\
\begin{split}
    \label{eq:lstm-output-gate}
    \bm{o}_t = \sigma(\bm{W}_{ox} \bm{x}_t + \bm{W}_{oh} \bm{h}_{t-1} + \bm{b}_o)
\end{split}\\
\begin{split}
    \label{eq:lstm-gate}
    \bm{g}_t = \text{tanh}(\bm{W}_{gx} \bm{x}_t + \bm{W}_{gh} \bm{h}_{t-1} + \bm{b}_g)
\end{split}\\
\begin{split}
    \label{eq:lstm-new-cell-state}
    \bm{c}_t = \bm{c}_{t-1} \odot \bm{f}_t + \bm{i}_t \odot \bm{g}_t
\end{split}\\
\begin{split}
    \label{eq:lstm-activation}
    \bm{h}_t = \text{tanh}(\bm{c}_t \odot \bm{o}_t)
\end{split}
\end{align}

\begin{figure}[t]
    \centering
    \includegraphics[width=\linewidth]{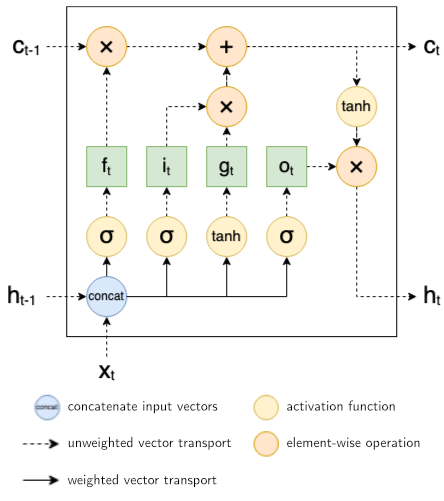}
    \caption{Schematic of the operations associated within an LSTM hidden unit. The computed internal state $\bm{c}_t$, and output $\bm{h}_t$ calculated at time step $t$ form the next inputs to the same cell at time step $t+1$ along with the next input vector $\bm{x}_{t+1}$ in the sequence.}
    \label{fig:LSTM-cell}
\end{figure}

The weight matrices associated with the inputs $\bm{x}_t$ and $\bm{h}_{t-1}$ destined for each gate are given by $\bm{W}_{fx}$, $\bm{W}_{fh}$, $\bm{W}_{ix}$, $\bm{W}_{ih}$, $\bm{W}_{ox}$, $\bm{W}_{oh}$, $\bm{W}_{gx}$ and $\bm{W}_{gh}$ and the bias vectors are given by $\bm{b}_f$, $\bm{b}_i$, $\bm{b}_o$ and $\bm{b}_g$. The $\odot$ operator denotes element-wise multiplication and $\sigma$ is an application of the sigmoid function. A schematic of the internal workings of an individual LSTM cell is given in \autoref{fig:LSTM-cell}.

For our experiments we took inspiration from \cite{Kong:cn} and designed our LSTM network using 2 connected layers, each containing 20 hidden LTSM cells followed by a single linear feed-forward layer. The loss function used for optimisation was a simple mean squared error. The Adam optimiser was used for training the network in all experiments using the recommended default hyperparameters in \cite{Kingma:2014us} combined with a fixed learning rate of 0.001 and a fixed batch size of 256 sequences.

\subsection{Training scenarios}

\begin{figure*}[]
    \centering
    \includegraphics[width=\textwidth]{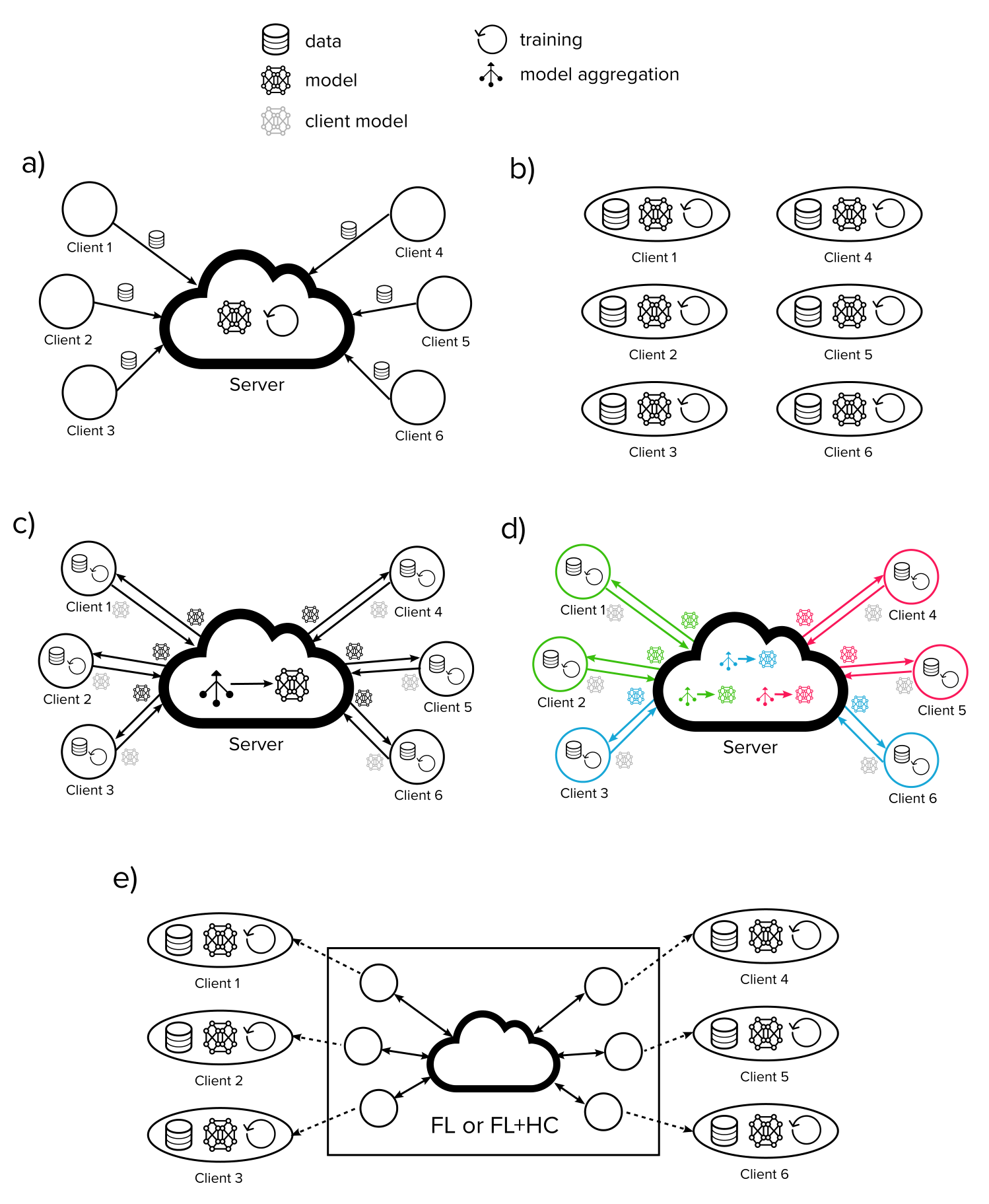}
    \caption{Diagrams detailing the the different training scenarios. a) Centralised learning - data is sent from clients to server, model and training is on the server. b) Localised training - data, model and training are isolated to each client. c) FL - data and training isolated to each client, model is aggregated from client updates at the server. d) FL+HC - after n rounds of FL, clients are clustered and model updates from each cluster are aggregated to specialised cluster models at the server (colours represent clusters). e) Local fine-tuning (LFT) - an extra step after either FL or FL+HC where training is isolated to each client starting with the model produced at the FL stage.}
    \label{fig:training-scenarios}
\end{figure*}

In order to test the effectiveness of applying FL to load forecasting, we provide benchmarks against centralised learning, local-only learning and various FL training scenarios. These different training approaches are summarised in \autoref{table:training-scenarios} and described diagrammatically in \autoref{fig:training-scenarios}.

Firstly we developed a non-distributed, centralised learning approach that is most commonly applied where the privacy of data is not a major concern during training. This approach pools individual household datasets together and training is conducted in a single location. This approach provides a baseline for what a single, joint forecasting model can achieve in a non-private setting. In this scenario, the same network parameters are used by all households at the inference stage. Under this centralised approach we train models for 500 epochs with early stopping based on the lowest error achieved on the validation set.

The most important benchmark we developed is a fully-private localised learning setting. All individual datasets remain private and unseen by other data owners under this scenario and the training procedure is isolated to each household. This approach results in unique forecasting models tailored to each household but cannot benefit from knowledge that could be embedded in data owned by other households. As per the centralised learning approach, training was conducted for a maximum of 500 epochs with early stopping.

\begin{table}[t]
\centering
\small
\caption{Summary of training scenarios, the forecasting models produced by each scenario and the privacy associated with each scenario.}
\begin{tabular}{ l l l }
\hline
Training scenario & Model & Private?\\
\hline
Centralised & Single/Joint & \ding{55}\\
Localised & Multiple/Specialised & \ding{51}\\
FL & Single/Joint & \ding{51} w.r.t. raw data\\
FL+HC & Multiple/Specialised & \ding{51} w.r.t. raw data\\
FL $\rightarrow$ LFT & Multiple/Specialised & \ding{51} w.r.t. raw data\\
FL+HC $\rightarrow$ LFT & Multiple/Specialised & \ding{51} w.r.t. raw data\\
\hline
\end{tabular}
\label{table:training-scenarios}
\end{table}

Any FL system needs to offer benefits above and beyond what can be achieved in the localised learning setting as this is already a fully private approach. In this paper we investigate how individual learners can benefit from patterns in energy usage from other households in the population. 

The goal of FL is the same as centralised learning - to learn a single, joint model that generalises well enough to provide accurate forecasts for all individual households. In FL however, the training data belonging to each household (or client in the parlance of FL) is not pooled as in centralised learning. Instead, the training data remains private to each local client. Whereas centralised learning seeks to optimise a global objective of the form: $\min f(w)$, FL optimises an objective as the finite sum of local objectives taking the form: $\min \frac{1}{m} \sum^m_{i=1} f_i(w)$.

Training proceeds via communication rounds, beginning with an initialised model state $w_t$ that is transmitted to a small set of clients $K$. Each client $k \in K$ computes an update $w^k_{t+1}$ to the model state based on their dataset by optimising a local forecasting objective $f_k(w_t)$. In practice this usually involves training just a few epochs on each client. Each client then transmits their update to a centralised server that aggregates the updates into a new model $w_{t+1}$. For our experiments we apply federated averaging (FedAvg) \cite{mcmahan2017communication} as the FL algorithm for aggregating client updates. FedAvg aggregates incoming client model updates via a data weighted average such that:

\begin{equation}
    w_{t+1} = \sum^K_{k=1} \frac{n_k}{n} w^k_{t+1}
\end{equation}

Here, $\frac{n_k}{n}$ represents the number of samples available to client $k$ compared to the total number of samples used for training in round $t$, thus determining the data-weighted contribution of client $k$. In the FL training scenario, model performance is affected by additional hyperparameters that we also test for:

\begin{itemize}
    \item fraction of clients participating in each communication round: 0.1, 0.2 \& 0.3
    \item Number of epochs of training on clients: 1, 3 \& 5
\end{itemize}

All FL training runs are capped at 500 communication rounds with early stopping based on the best average validation set performance across all clients.

Models trained via FL have been shown to suffer under non-IID distributions. As we have explored, the individual household datasets exhibit differing data distributions due to the varied ways in which household occupants consume energy. As such, we investigate the use of a modification of FedAvg known as FL+HC \cite{Briggs:it} (our previous work) that incorporates a clustering step into the FL protocol. In FL, the local objectives are expected to approximate the global objective, however if data among clients is distributed non-IID, this expectation over data available to client $D_k$ is not valid: $\mathbb{E}_{D_k}[f_k(w)] \neq f(w)$. In FL+HC, clients are assigned to a cluster $c \in C$, where the goal is to train a specialised model $f_c$ tailored to clients that share a similar data distribution. We use client updates as a proxy for client similarity in order to preserve the privacy of the raw training data. A hierarchical clustering algorithm is run at communication round $n$ taking as input the weight updates from all clients. Clients that produce similar updates are clustered together and further training via FL proceeds for each cluster in isolation. For a good clustering under FL+HC, the expectation of local objectives (clients) assigned to a cluster $c$ approximates the cluster objective:

\begin{equation}
	\forall{c \in C}, \mathbb{E}_{D_k}[f_k(w)] = f_c(w) \text{ where } k \in c
\end{equation}

FL+HC introduces more hyperparameters to control the clustering process which we test for:

\begin{itemize}
    \item clustering distance threshold: 0.8, 1.4 \& 2.0
    \item hierarchical clustering linkage mechanism: ward, average, complete \& single
    \item number of rounds of FL prior to clustering step $n$: 3, 5 \& 10
\end{itemize}

To keep the number of permutations of experiments for the FL+HC scenario manageable, we fix the client fraction at 0.1, the number of epochs to 3 and exclusively use the Euclidean (L2) clustering distance metric. All FL+HC training runs are capped at 200 communication rounds with early stopping based on the best average validation set performance per cluster.

Finally, we also test scenarios where the models produced by FL and FL+HC are further fine-tuned on the local clients for a small number of epochs (a process known as personalisation). These local fine-tuning (LFT) scenarios are termed FL $\rightarrow$ LFT and FL+HC $\rightarrow$ LFT. We test whether personalisation produces more accurate, highly specialised models for each household that also builds on the wisdom of the private data of other households. The fine-tuning step is limited to 25 epochs on all clients with early stopping based on the best validation set performance.

\subsection{Running the experiments}
All experiments are run in the PyTorch deep learning framework \cite{NEURIPS2019_9015} on the Google Cloud Computing (GCP) platform using a single NVIDIA Tesla K80 GPU attached to 30GB of memory, a 128GB SSD and 8 virtual CPUs on an Intel Xeon processor (n1-standard-8 GCP machine type). The distributed training scenarios are all simulated on a single machine.
\section{Results \& Discussion}
\label{sec:results-and-discussion}

\subsection{Forecasting performance}
\begin{table*}[t]
\centering
\small
\caption{Forecasting error (RMSE) values for the 6 datasets and 6 training scenarios. 'Mean' and 'Best' columns show percentage difference compared to fully-private localised learning}
\begin{tabular}{ l c c c c c c l l }
\toprule
& \multicolumn{6}{c}{Dataset} & &\\
\cmidrule(lr){2-7} & \multicolumn{3}{c}{incl. weather} & \multicolumn{3}{c}{not incl. weather}&\\
\cmidrule(lr){2-4} \cmidrule(lr){5-7}
Training scenario & 6 steps & 12 steps & 24 steps & 6 steps & 12 steps & 24 steps & \multicolumn{1}{c}{mean} & \multicolumn{1}{c}{best}\\
\midrule  
Centralised             & 0.0210 & 0.0207 & 0.0198 & 0.0210 & 0.0208 & 0.0198 & 0.0205 \hfill (-4.8\%) & 0.0198 \hfill (-8.0\%)\\
Localised               & 0.0198 & 0.0183 & 0.0201 & 0.0201 & 0.0188 & 0.0202 & 0.0196 \hfill (---) & 0.0183 \hfill (---)\\
FL                      & 0.0219 & 0.0202 & 0.0207 & 0.0219 & 0.0205 & 0.0210 & 0.0210 \hfill (-7.6\%) & 0.0202 \hfill (-10.5\%)\\
FL+HC                   & 0.0209 & 0.0189 & 0.0208 & 0.0214 & 0.0198 & 0.0210 & 0.0205 \hfill (-4.7\%) & 0.0189 \hfill (-3.5\%)\\
FL $\rightarrow$ LFT    & 0.0194 & 0.0177 & 0.0192	& 0.0195 & \bf{0.0174} & 0.0192 & 0.0187 \hfill (+4.3\%) & 0.0174 \hfill (+4.9\%)\\
FL+HC $\rightarrow$ LFT & \bf{0.0193} & \bf{0.0176} & \bf{0.0192} & \bf{0.0193} & 0.0175 & \bf{0.0192} & 0.0187 \hfill (+4.5\%) & 0.0175 \hfill (+4.5\%)\\
\bottomrule
\end{tabular}
\label{table:results-performance}
\end{table*}

To understand how the different training approaches perform compared to one another, we report the RMSE achieved on the test set for each of the 6 datasets, $\bm{S}_{K=6,\text{+weather}}$, $\bm{S}_{K=12,\text{+weather}}$, $\bm{S}_{K=24,\text{+weather}}$, $\bm{S}_{K=6,\text{-weather}}$, $\bm{S}_{K=12,\text{-weather}}$ and $\bm{S}_{K=24,\text{-weather}}$. The RMSE reported is an average over all clients (in distributed approaches) or over all sequences (in the centralised approach).
For experiments that involve tuning hyperparameters (namely those that involve FL), we report the test set RMSE for the best performing model based on the lowest error on the validation set. We also report the mean RMSE and lowest RMSE over all datasets for each training approach along with a percentage difference to compare with the fully private localised approach. Model performance results are detailed in \autoref{table:results-performance}. 

In the centralised approach, the training procedure has access to all sequences pooled from across the individual household datasets. Therefore model performance might be expected to be relatively high compared to the other approaches where there is much less data to learn from. Conversely, we show that average model performance in the centralised approach is actually 4.8\% worse than the localised approach and the best centralised model is 8.0\% worse than the best localised model. This is somewhat surprising given that the localised models only have access to $1/100$ the amount of data. This implies that the centralised models (and possibly single, joint models in general) struggle to capture individual household behaviours in energy usage and/or suffer from trying to optimise for competing objectives. Larger models might allow for learning more individual behaviours but as data has to be gathered into a single location, the privacy risk to energy consumers is by far the highest in this training approach. The 24-step sequence (1 whole of day of prior readings) provides the model with the most information with which to make a prediction, resulting in the lowest RMSE in the centralised approach (followed by the 12-step, then 6-step).

In the localised learning approach, a model for each household is trained in isolation using only the data available to that household. Model performance is exceptionally good in this approach and the simple LSTM architecture is sufficient to learn more nuanced energy demand behaviours unique to each household. This approach represents a fully private setting in that nothing is shared between households. Datasets formed around 12-step sequences result in models that significantly outperform 6-step and 24-step sequence datasets in this approach. We see a similar pattern for the remaining training approaches, suggesting that a 12-hour time window is optimal for local learners to most accurately predict future energy demand.

In the FL approach, only a fraction of clients are selected for each round of training (and each client trains on its local data set in isolation for a small number of epochs). Additionally a single, joint model is being co-trained by these selected clients when model updates are aggregated. As such, we see that the RMSE for FL models suffers in the same way as centralised models when we compare to the localised approach. Additionally, as FL has been shown to perform sub-optimally in cases where the training data is non-IID as is the case with the individual household datasets, the RMSE suffers even more so than in the centralised learning approach. Compared with localised learning the average model performance in the FL approach was 7.6\% worse with the best FL model significantly worse (10.5\% higher RMSE) than the best localised learning model.

The FL+HC approach produces specialised models for a number of clusters of clients that can more specifically tailor forecasts for groups of households that provide similar model updates (a proxy for similar underlying energy demand distributions across clients). As such, the average RMSE of clients is no longer tied to a single, joint model as in the FL training approach, but rather to a specialised cluster model. In the average case across the 6 datasets, FL+HC produces models 4.7\% worse than the localised approach - comparable to centralised learning. However, the best model trained with FL+HC significantly outperforms models trained with FL or centralised learning but remains 3.5\% worse than localised learning. Although the FL approaches (FL and FL+HC) do occasionally produce a slightly better model than the centralised training scenario, the mean RMSE across all datasets shows that on average FL performance is degraded compared to centralised learning, consistent with the findings of most previous FL literature.

\begin{table*}[t]
\centering
\small
\caption{Computational efficiency (measured in millions of samples required to train top performing models) for the 6 datasets and 6 training scenarios. 'Mean' and 'Best' columns show savings in computation compared to fully-private localised learning.}
\begin{tabular}{ l r r r r r r r l r l }
\toprule
& \multicolumn{6}{c}{Dataset} & &\\
\cmidrule(lr){2-7} & \multicolumn{3}{c}{incl. weather} & \multicolumn{3}{c}{not incl. weather}&\\
\cmidrule(lr){2-4} \cmidrule(lr){5-7} Training scenario & 6 steps & 12 steps & 24 steps & 6 steps & 12 steps & 24 steps & \multicolumn{2}{c}{mean} & \multicolumn{2}{c}{best*}\\
\midrule  
Centralised & 78.9 & 19.5 & 21.2 & 15.6 & 26.6 & 57.7 & 36.6 & (1.9x) & 57.7 & (1.4x)\\
Localised & 71.5 & 79.5 & 52.7 & 94.9 & 67.1 & 60.5 & 71.0 & (---) & 79.5 & (---)\\
FL & 266.3 & 290.7 & 384.4 & 238.1 & 424.3 & 455.3 & 343.2 & (0.2x) & 290.7 & (0.3x)\\
FL+HC & 6.3 & 6.1 & 6.8 & 7.3 & 7.0 & 0.1 & 5.6 & (12.7x) & 6.1 & (13.1x)\\
FL $\rightarrow$ LFT & 268.4 & 292.5 & 388.0 & 239.2 & 429.6 & 458.2 & 346.0 & (0.2x) & 429.6 & (0.2x)\\
FL+HC $\rightarrow$ LFT & 7.2 & 6.9 & 7.9 & 8.0 & 8.7 & 1.2 & 6.7 & (10.7x) & 8.7 & (9.1x)\\
\bottomrule
\multicolumn{11}{l}{\footnotesize{}}\\
\multicolumn{11}{l}{\parbox{15cm}{\footnotesize{* the best savings are calculated by comparing the best performing model for each training scenario to the best performing localised model, as reported in \autoref{table:results-performance}.}}}\\
\end{tabular}
\label{table:results-computation}
\end{table*}

Although the base models trained with FL and FL+HC show a higher RMSE than those trained with fully private localised learning, we now show how the situation can be improved if we treat FL or FL+HC as a pre-training task to be followed by further fine-tuning on the local clients in isolation. In the FL $\rightarrow$ LFT approach, we use each joint model trained under FL on each of the 6 datasets and perform a small amount of further training per client to produced highly specialised models. The trained parameters of the base models serve as a good initialisation point for rapid training on the clients which often converge within just a few epochs of fine-tuning. These personalised models exhibit a lower RMSE than the other approaches across all datasets. On average FL $\rightarrow$ LFT produces models with a 4.5\% lower RMSE than localised training with the best model performing 4.9\% better than the best localised model. In FL+HC $\rightarrow$ LFT approach, clients initialise their personalised models from the specialised model trained within the cluster each client belongs to. This approach produces similarly performing models (4.5\% better than the average and best localised model). These personalisation approaches clearly show that local models can benefit by learning from the energy demand patterns of other users. FL allows for energy consumers to contribute to the shared learning task whilst retaining the privacy of their raw consumption data prior to privately fine-tuning their own models to produce more accurate forecasts.

The datasets that included weather features ($\bm{S}_{K=6,\text{+weather}}$, $\bm{S}_{K=12,\text{+weather}}$ and $\bm{S}_{K=24,\text{+weather}}$) show a small improvement in model performance in almost all scenarios compared to datasets without such features. We would therefore recommend that an load forecasting system should make use of weather related features if possible as these indicators can help the model to make better predictions on the whole.

\subsection{Computational efficiency}
In addition to measuring the accuracy of forecasts produced by the various training approaches, we note the computational efficiency via the number of samples passing through the optimiser during training. In this sense we can understand how many data samples are required to train the best performing model for each training scenario/dataset. Fewer training samples corresponds with models that can be trained with less computational effort - a desirable characteristic if the forecasting task is to be run at the network edge on low-compute smart meter devices in the homes of energy consumers. In FL it is also very desireable to reduce the amount of communication during training which would otherwise require large amounts of bandwidth. In a practical implementation of FL, communicating large models between household smart meters and entities coordinating the model training could become a bottleneck in the learning process. Any measures to reduce the total number of communication rounds will be beneficial, therefore we present a study of the computational efficiency of each training method within this section. We provide results for computational efficiency (measured in millions of samples) in \autoref{table:results-computation}. We also benchmark each method against the fully private localised case, reporting average savings in computation and the savings for the best performing models for each training scenario. In all scenarios where multiple specialised models are trained, we report the total number of samples required to train all models (be they clustered or individual to each client).

Although non-private we noted that 36.6 million samples were required on average to train the single, joint centralised model. The fewest samples (15.6 million) were required for the model trained using the $\bm{S}_{K=6,\text{-weather}}$ dataset. Training the best performing centralised model required 1.4x fewer samples than were required to train all the localised models. Training the individual localised models required 71.0 million samples on average and 79.5 million samples for the best performing model using the $\bm{S}_{K=12,\text{+weather}}$ dataset.

Under the FL training scenario, many hundreds of communication rounds were required to reach the minimum training loss before overfitting. Although only a fraction of clients are selected during each communication round, training proceeds on each client for a number of epochs. These factors lead to a significant amount of computation in total over the whole training operation. On average 5x more computation is required to train the FL models vs the localised models. The best performing FL model required nearly 4x more computation to train vs the best localised models.

The FL+HC training scenario only trains a single, joint model for a small number of rounds prior to producing specialised models at the clustering step. Post-clustering, each specialised model is exclusively trained on the cluster's subset of clients. This has the effect of drastically reducing the amount of computation required to train each model. In \autoref{table:results-computation} We report the total number of samples required to train the FL+HC models across all the clusters. There is a drastic saving in computation in this training scenario as FL+HC strikes a good balance between learning from all clients initially to  produce specialised models that are quick to train. On average FL+HC requires 12.7x fewer samples to train models vs localised training and 13.1x fewer samples to train the best performing models.

Where FL and FL+HC are followed by a fine tuning step, only a few epochs of training are required to produce the best performing personalised models. Therefore very little extra computation is required to fine-tune. As we showed earlier, these personalised models exhibit the lowest error of all the models we tested and FL+HC $\rightarrow$ LFT in particular produces low error models with $\sim$10x reduction in computation compared to localised training.

\section{Conclusion}
\label{sec:conclusion}

In this paper, we explored the use of FL for the purpose of private load forecasting using an LSTM network. We compared our results with benchmarks - a non-private centralised training approach and a fully private localised learning approach. Additionally we investigated the use of FL+HC - a clustered variant of FL shown to perform well on non-IID data. We determined that FL approaches can outperform centralised learning but perform worse than localised learning. We presented favourable results however, when a personalisation step is applied to the models trained by FL and FL+HC. In this case model performance can be improved by up to 5\% compared to localised learning while still retaining the privacy of the raw energy consumption data. We also reported the computational efficiency of the various training methods, concluding that FL+HC and FL+HC followed by fine-tuning result in vast computational savings (on the order of 10x reduction) in the number of samples required to train the best models. Finally we provide some brief advice on aggregation of predictions after the training procedure to inform the design of a complete privacy-preserving training/inference framework for load forecasting.

\section*{Acknowledgments}
This work is partly supported by the SEND project (grant ref. 32R16P00706) funded by ERDF and BEIS as well as EPSRC EnergyREV (EP/S031863/1).


\ifCLASSOPTIONcaptionsoff
  \newpage
\fi



\bibliographystyle{IEEEtran}
\bibliography{IEEEabrv,../main}
\end{document}